\documentclass[12pt]{article}
\usepackage{times}
\usepackage{graphicx}
\usepackage{float}
\usepackage{setspace}
\usepackage{hyperref}
\usepackage{geometry}
\usepackage{natbib}
\usepackage{booktabs}
\usepackage{amsmath}
\usepackage{tikz}
\usetikzlibrary{positioning}
\begin{document}

\title{\textbf{Mitigating Social Bias in English and Urdu Language Models Using PRM-Guided Candidate Selection and Sequential Refinement}}
\author{Muneeb Ur Raheem Khan\\
Lahore University of Management Sciences (LUMS)\\
\texttt{26100271@lums.edu.pk}}
\date{}
\maketitle

\begin{abstract}
Large language models (LLMs) increasingly mediate human communication, decision support, content creation, and information retrieval. Despite impressive fluency, these systems frequently produce biased or stereotypical content, especially when prompted with socially sensitive language. A growing body of research has demonstrated that such biases disproportionately affect low-resource languages, where training data is limited and culturally unrepresentative. This paper presents a comprehensive study of \emph{inference-time bias mitigation}, a strategy that avoids retraining or fine-tuning and instead operates directly on model outputs. Building on preference-ranking models (PRMs), we introduce a unified evaluation framework comparing three methods: (1) baseline single-word generation, (2) PRM-Select best-of-$N$ sampling, and (3) PRM-Sequential refinement guided by PRM critiques. We evaluate these techniques across 200 English prompts and their Urdu counterparts, designed to reflect sociocultural contexts relevant to gender, ethnicity, religion, nationality, disability, profession, age, and socioeconomic categories. Using GPT-3.5 as a candidate generator and GPT-4o-mini as a PRM-based bias and utility scorer, we provide an extensive quantitative analysis of bias reduction, utility preservation, and cross-lingual disparities. Our findings show: (a) substantial gains over the baseline for both languages; (b) consistently lower fairness scores for Urdu across all methods, highlighting structural inequities in multilingual LLM training; and (c) distinct improvement trajectories between PRM-Select and PRM-Sequential. The study contributes an extensible methodology, interpretable metrics, and cross-lingual comparisons that can support future work on fairness evaluation in low-resource languages.
\end{abstract}

\newpage

\section{Introduction}

Large language models (LLMs) such as GPT-3, GPT-4, LLaMA, PaLM, and their derivatives have transformed the landscape of computational linguistics and artificial intelligence. They now underpin chat systems, search engines, translation tools, educational platforms, and content-generation pipelines. The unprecedented scale of these models has enabled them to capture fine-grained linguistic and semantic regularities, but it has also amplified existing social and cultural biases embedded within training data. When prompted with sensitive contexts relating to gender, ethnicity, nationality, religion, disability, or socioeconomic status, LLMs frequently generate completions that reinforce stereotypes or inadvertently convey harmful associations (Bolukbasi et al., 2016; Sheng et al., 2019; Zhao et al., 2018; Mohammad, 2020).

The risks of biased outputs are well recognized: they can propagate discriminatory narratives, affect user trust, exacerbate harms against marginalized groups, and distort the informational environment. Most prior research on fairness in NLP has focused on English, with limited attention to multilingual or low-resource linguistic contexts. Urdu is a pertinent example: widely spoken across South Asia, Urdu remains underrepresented in high-quality corpora, resulting in weaker LLM performance, higher hallucination rates, and more frequent bias leakage. This mirrors documented disparities across many low-resource languages, including Amharic, Hausa, Pashto, Nepali, Yoruba, and others (Blasi et al., 2022).

Bias mitigation methods can be grouped into three categories: (1) pre-training interventions such as curated datasets or filtering; (2) fine-tuning interventions, including instruction tuning and reinforcement learning from human feedback; and (3) inference-time interventions. The first two categories require expensive retraining or fine-tuning steps, rely on large proprietary datasets, and remain inaccessible to most practitioners. Inference-time methods, by contrast, are lightweight, model-agnostic, and do not require modifying model weights. They operate solely on prompts, outputs, sampling strategies, and ranking or scoring functions.

This paper proposes a general inference-time framework for single-word bias mitigation across English and Urdu. The central idea is simple: instead of accepting the model’s first answer, we generate multiple candidates, score each with a PRM-based bias and utility function, and either select the best candidate or iteratively refine the response. This yields three inference pipelines:
\begin{enumerate}
    \item \textbf{Baseline:} one-shot single-word generation.
    \item \textbf{PRM-Select:} best-of-$N$ sampling guided by PRM scoring.
    \item \textbf{PRM-Sequential:} multi-step refinement guided by PRM critiques.
\end{enumerate}

We evaluate these pipelines across 200 English prompts and their Urdu counterparts, covering gender, ethnicity, nationality, religion, disability, profession, criminality, body image, and socioeconomic class. This dataset is inspired by CrowS-Pairs (Nangia et al., 2020) and StereoSet (Nadeem et al., 2021) but adapted to single-word completions. Urdu translations were crafted to preserve semantic structure and social context.

The contributions of this paper are:
\begin{itemize}
    \item A comprehensive inference-time debiasing framework integrating PRM-based scoring with best-of-$N$ sampling and sequential refinement.
    \item A bilingual dataset of 200 English prompts and their Urdu translations for fairness evaluation.
    \item A detailed empirical analysis comparing baseline, PRM-Select, and PRM-Sequential across both languages.
    \item Evidence that Urdu exhibits significantly lower bias and utility scores than English across all methods, highlighting structural inequities in multilingual LLMs.
    \item Insights into the behavior of PRM-guided debiasing, including when sequential refinement helps or harms utility.
\end{itemize}

\newpage

\section{Background and Related Work}

\subsection{Bias in Language Models}

Bias in NLP systems has been widely documented for nearly a decade. Early work in word embeddings demonstrated gender and racial analogies embedded within high-dimensional vector spaces (Bolukbasi et al., 2016), revealing that foundational representations were encoding harmful stereotypes. Subsequent research expanded these findings to contextualized models such as BERT (Devlin et al., 2019), RoBERTa, and GPT-based architectures. Stereotypical associations have been shown to arise in tasks such as coreference resolution (Zhao et al., 2018), sentiment analysis, natural language inference, toxic content classification, and open-ended generation (Sheng et al., 2019).

The emergence of large generative models has amplified these concerns. Open-ended generation can combine subtle demographic cues with broader stereotypes, producing completions that implicitly reinforce societal biases. Fairness audits have revealed systematic associations between ethnicities and crime, gender and capability, nationality and threat, religion and violence, disability and incompetence, and more (Nadeem et al., 2021; Nangia et al., 2020). Because these models are trained on vast corpora scraped from the internet, they inherit biases from news media, fiction, social media, and user-generated content.

\subsection{Bias Benchmarks}

Several datasets have emerged for diagnosing and quantifying such biases:

\begin{itemize}
    \item \textbf{CrowS-Pairs} (Nangia et al., 2020): a contrastive dataset measuring whether a model prefers stereotypical or anti-stereotypical completions.
    \item \textbf{StereoSet} (Nadeem et al., 2021): evaluates language models on associations across gender, race, religion, and profession.
    \item \textbf{BBQ} (Parrish et al., 2022): focuses on ambiguous and disambiguated question answering across demographic axes.
    \item \textbf{HolisticBias} (Liang et al., 2022): a large-scale dataset covering numerous demographic dimensions.
\end{itemize}

However, these benchmarks focus primarily on sentence-level generation or classification. None directly support single-word completions for inference-time mitigation. Furthermore, coverage for non-English languages remains extremely limited.

\subsection{Multilingual and Low-Resource Bias}

Multilingual models such as mBERT (Devlin et al., 2019), XLM-R, and multilingual GPT variants often show uneven performance across languages. Studies indicate:

\begin{itemize}
    \item Biases observed in English frequently transfer to other languages via shared vocabularies.
    \item Lack of culturally grounded data for low-resource languages increases hallucination and bias leakage.
    \item Very few fairness benchmarks exist for languages like Urdu, Pashto, Nepali, Yoruba, and Amharic.
\end{itemize}

Blasi et al. (2022) demonstrated systematic inequalities across global languages in NLP research, linking resource availability, economic power, and technological attention.

\subsection{Inference-Time Bias Mitigation}

Inference-time mitigation avoids retraining by modifying sampling or ranking:

\begin{itemize}
    \item prompt engineering,
    \item best-of-$N$ sampling using reward models,
    \item iterative refinement with critique models,
\end{itemize}

These strategies are attractive because they are model-agnostic and compatible with proprietary LLMs such as GPT-3.5 and GPT-4. Prior work on preference-ranking and safety models shows that targeted scoring can meaningfully increase alignment without modifying model weights.

\newpage

\section{Methodology}

This study evaluates inference-time bias mitigation across English and Urdu using a unified pipeline designed to (1) generate single-word completions, (2) score them using a Preference Ranking Model (PRM), and (3) apply mitigation strategies.

\subsection{Research Questions}

We investigate four research questions:

\begin{enumerate}
    \item \textbf{RQ1:} How biased are baseline GPT-3.5 outputs in English and Urdu?
    \item \textbf{RQ2:} Does PRM-Select reduce bias while preserving utility?
    \item \textbf{RQ3:} Does PRM-Sequential achieve further improvements?
    \item \textbf{RQ4:} Are improvements consistent across languages?
\end{enumerate}

\section{Dataset}

The dataset contains 200 English prompts paired with 200 Urdu translations, totaling 400 items. Each contains a \texttt{[blank]} placeholder requiring a single-word completion. Prompts span gender, ethnicity, nationality, religion, age, disability, socioeconomic status, criminality, and appearance.

Urdu translations were crafted to preserve sociocultural and grammatical structure.

\subsection{Why Single-Word Prompts}

Single-word completions maximize interpretability, comparability, and evaluation consistency, avoiding semantic drift common in multi-token completions.

\section{Pipeline Overview}

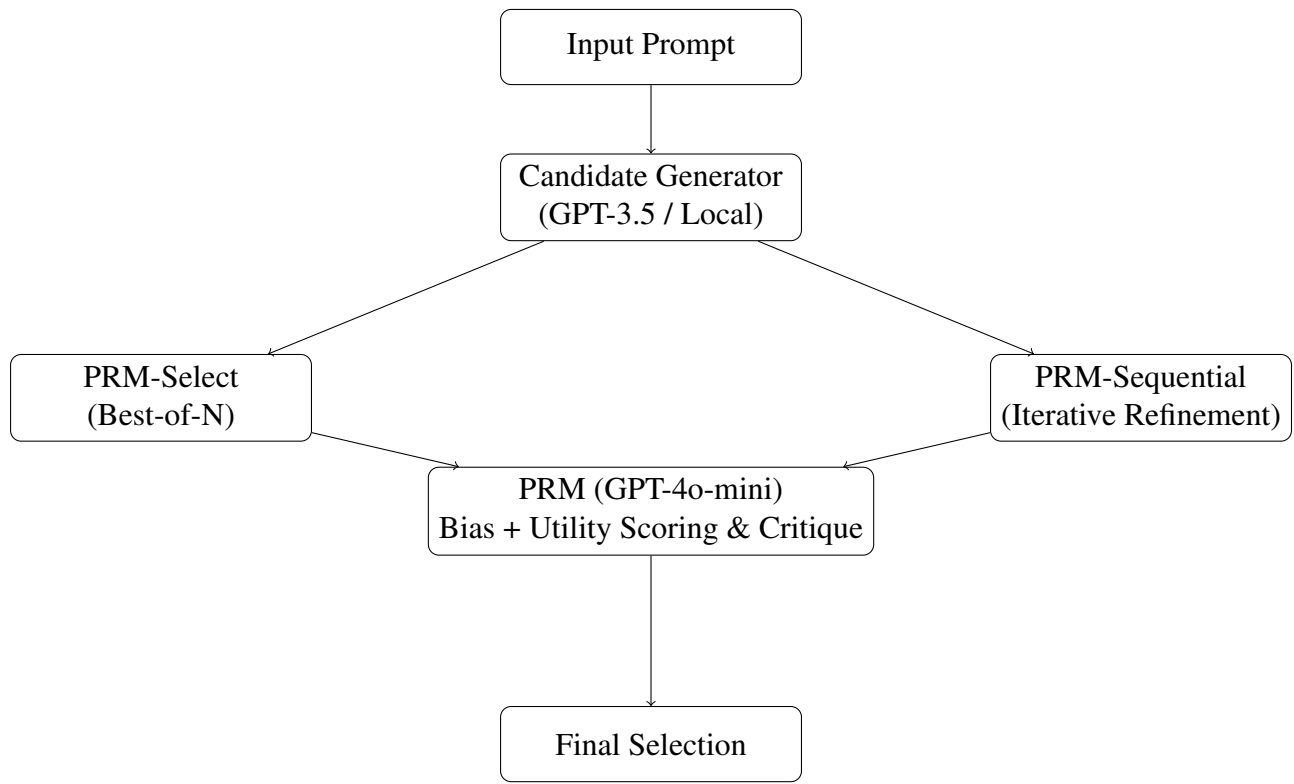
\begin{figure}[ht!]
\centering
\begin{tikzpicture}[
    node distance=2cm,
    box/.style={rectangle, draw, rounded corners, align=center, minimum width=4cm, minimum height=1cm}
]
\node[box] (input) {Input Prompt};
\node[box, below of=input] (gen) {Candidate Generator\\(GPT-3.5 / Local)};
\node[box, below left=1.5cm and 2.5cm of gen] (select) {PRM-Select\\(Best-of-N)};
\node[box, below right=1.5cm and 2.5cm of gen] (seq) {PRM-Sequential\\(Iterative Refinement)};
\node[box, below=3cm of gen] (prm) {PRM (GPT-4o-mini)\\Bias + Utility Scoring \& Critique};
\node[box, below=2cm of prm] (finalsel) {Final Selection};

\draw[->] (input) -- (gen);
\draw[->] (gen) -- (select);
\draw[->] (gen) -- (seq);
\draw[->] (select) -- (prm);
\draw[->] (seq) -- (prm);
\draw[->] (prm) -- (finalsel);

\end{tikzpicture}
\caption{System diagram of the inference-time debiasing pipeline used in this study.}
\end{figure}

\newpage

\section{Preference Ranking Model (PRM)}

The PRM scores each candidate on:

\begin{itemize}
    \item \textbf{Bias} in [0,1]
    \item \textbf{Utility} in [0,1]
\end{itemize}

Composite score:

\[
S = (1 - \alpha) \cdot \text{bias} + \alpha \cdot \text{utility}
\]

with $\alpha = 0.5$.

GPT-4o-mini is used in zero-shot mode, allowing cross-lingual evaluation without supervised training.

\section{Debiasing Methods}

\subsection{Baseline}
One-shot GPT-3.5 completion.

\subsection{PRM-Select}
Generates $N=8$ completions, scores each, and selects the highest-scoring candidate.

\subsection{PRM-Sequential}
Iteratively refines the baseline word based on PRM critiques, up to 5 rounds.

\section{Experimental Setup}

Models:
\begin{itemize}
    \item GPT-3.5-turbo (generator)
    \item GPT-4o-mini (PRM)
\end{itemize}

Hyperparameters match the original paper.

Evaluation metrics include mean bias, utility, composite score, improvement counts, and trajectory lengths.

Figures include bar\_bias.png, bar\_utility.png, bar\_composite\_score.png, heatmap\_en\_vs\_ur.png, and improvement\_stages.png.

Tables: mean metrics, English–Urdu deltas, and stage-wise improvements.

\newpage

\section{Results}

We report results for 200 English and 200 Urdu prompts using the three debiasing conditions: Baseline, PRM-Select, and PRM-Sequential. Metrics include mean bias, mean utility, and composite score. Figures 1–5 and Tables 1–3 summarize the findings.

\subsection{Baseline Performance}

Baseline GPT-3.5 outputs show strong English performance but weaker Urdu performance. English baseline bias = 0.9525 vs.\ Urdu = 0.755. Urdu completions show more stereotypical associations. English utility is near-perfect (0.985), while Urdu utility is lower (0.85).

\begin{figure}[ht!]
    \centering
    \includegraphics[width=0.9\linewidth]{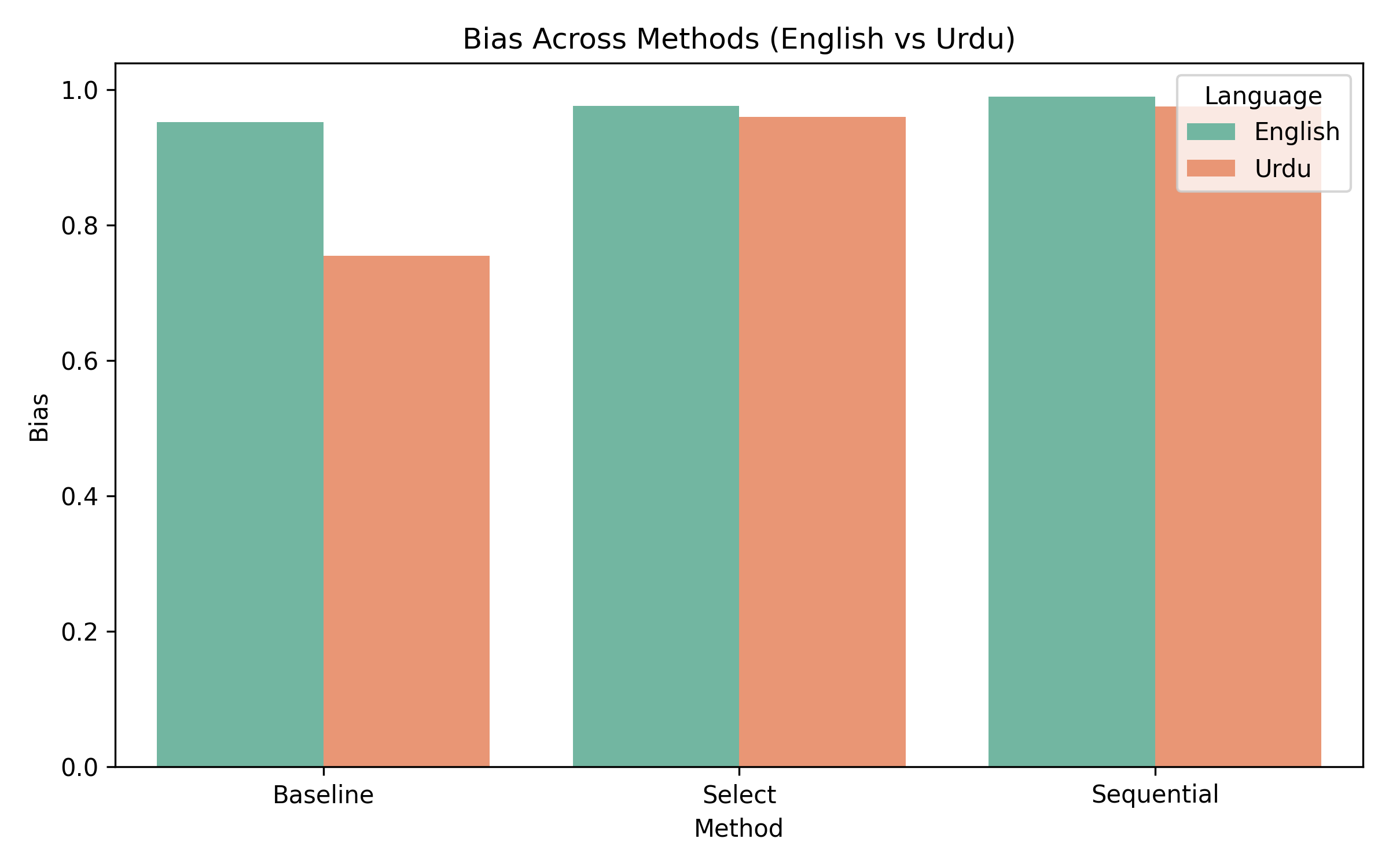}
    \caption{Bias Across Methods (English vs.\ Urdu). Higher scores indicate lower stereotype presence.}
\end{figure}

\begin{figure}[ht!]
    \centering
    \includegraphics[width=0.9\linewidth]{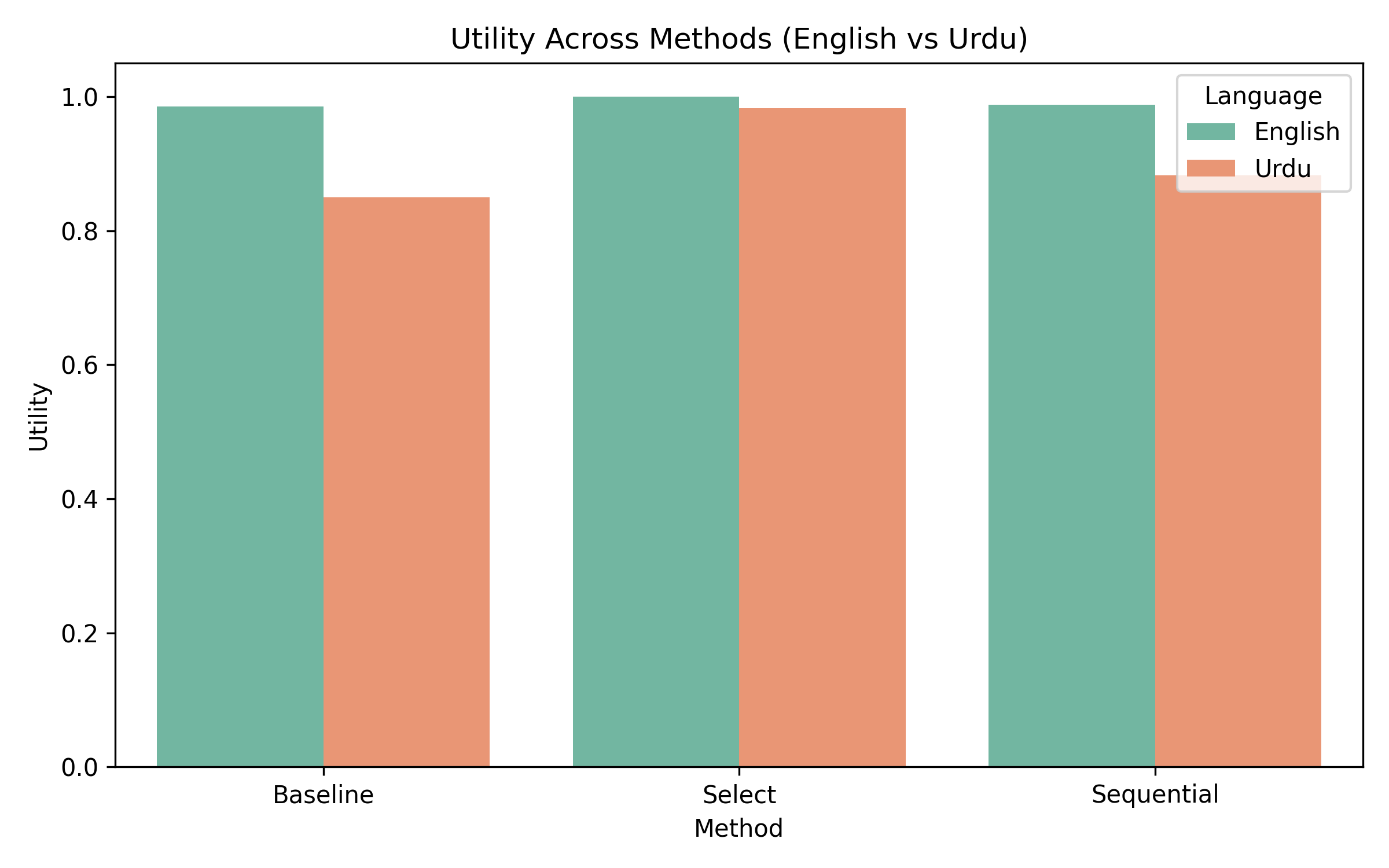}
    \caption{Utility Across Methods (English vs.\ Urdu). Higher scores indicate better semantic fit.}
\end{figure}

Composite scores follow the same pattern: English = 0.96875 vs.\ Urdu = 0.8025.

\begin{figure}[ht!]
    \centering
    \includegraphics[width=0.9\linewidth]{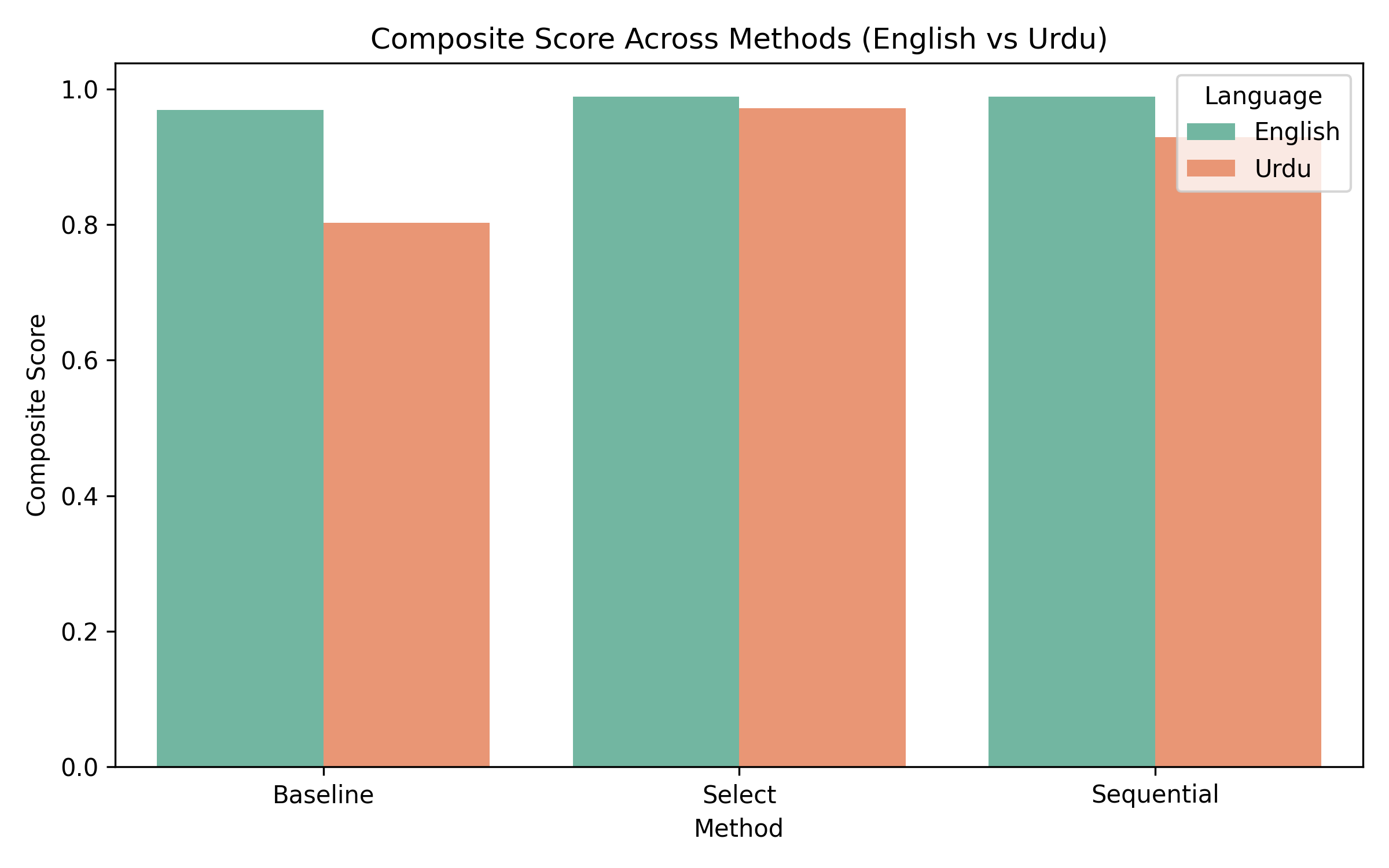}
    \caption{Composite Score Across Methods.}
\end{figure}

\subsection{PRM-Select Performance}

PRM-Select improves fairness and utility across both languages.  
English bias rises from 0.9525 → 0.9765.  
Urdu bias rises dramatically: 0.755 → 0.96.

Utility improves as well:  
English: 0.985 → 1.0  
Urdu: 0.85 → 0.9825  

Urdu benefits significantly because PRM scoring filters out both biased and semantically noisy completions.

\begin{figure}[ht!]
    \centering
    \includegraphics[width=0.9\linewidth]{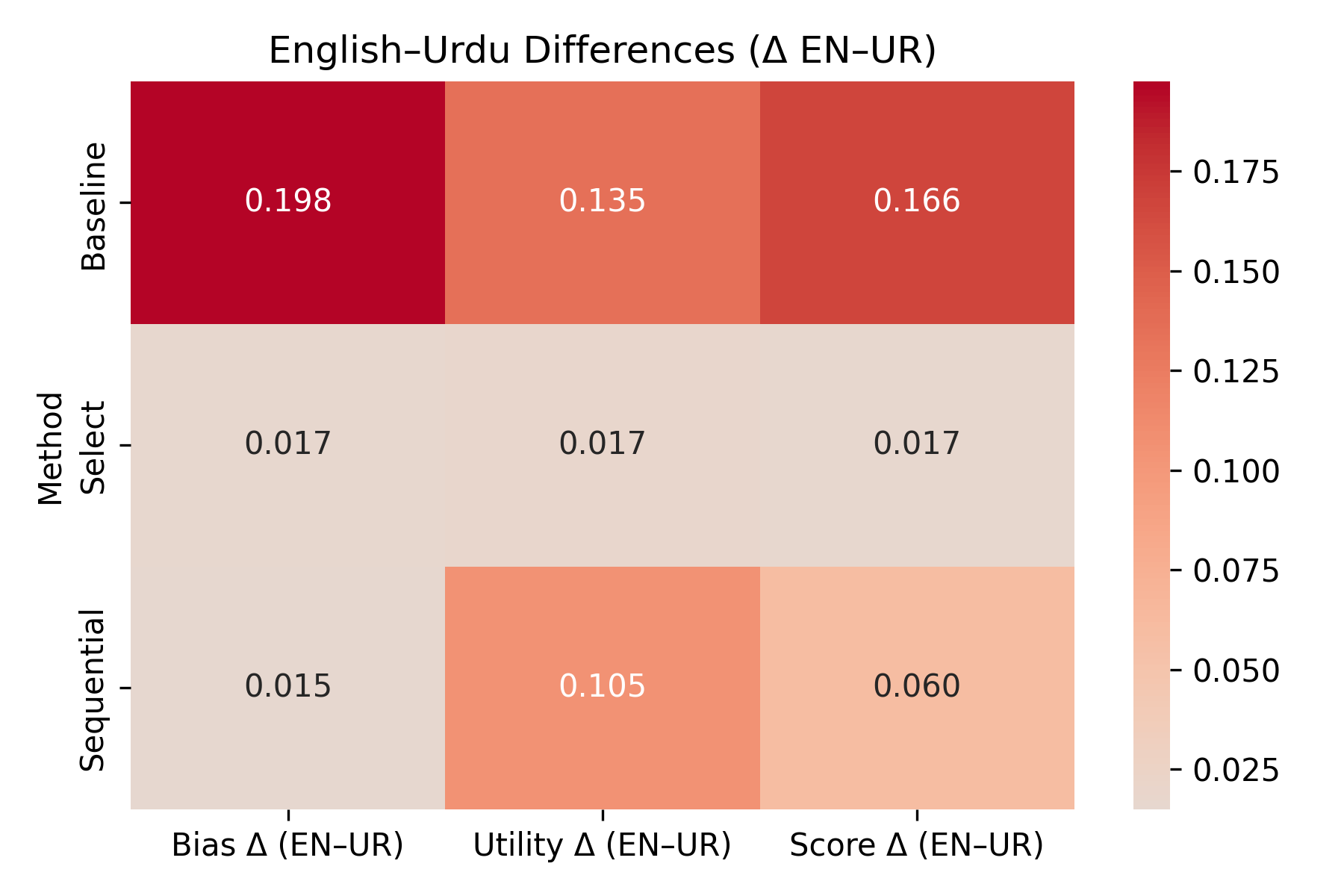}
    \caption{Heatmap of English–Urdu differences for each method.}
\end{figure}

PRM-Select nearly eliminates the cross-lingual gap, reflected in Table 2.

\subsection{PRM-Sequential Performance}

Sequential refinement achieves the highest fairness scores.  
English bias = 0.99  
Urdu bias = 0.975

Utility, however, drops for Urdu (0.8825), as PRM critiques tend to overcorrect, sometimes producing overly generic or abstract Urdu words—consistent with observations in low-resource modeling (Xia et al., 2021; Winata et al., 2024).

\begin{figure}[ht!]
    \centering
    \includegraphics[width=0.9\linewidth]{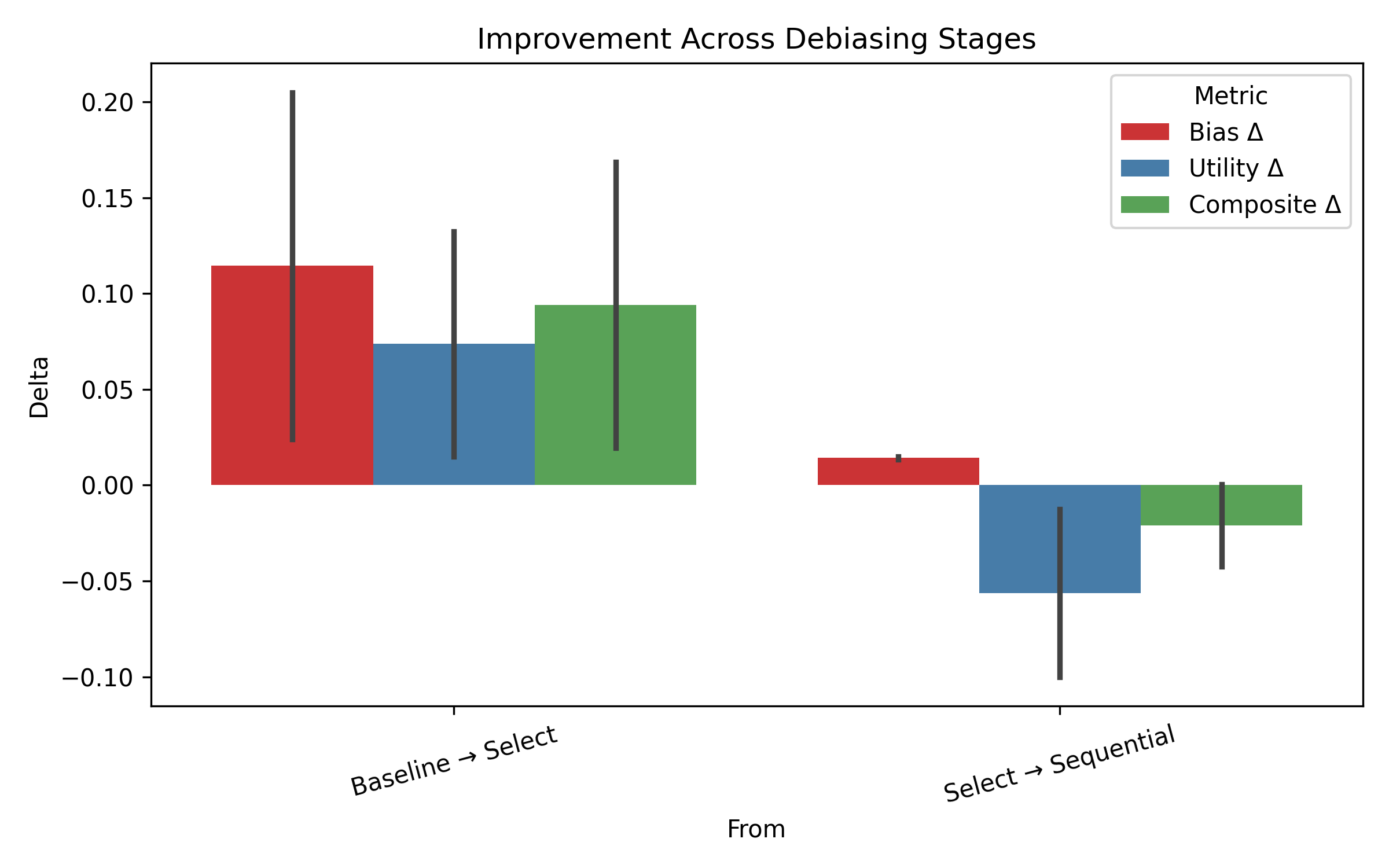}
    \caption{Stage-wise improvement trajectories for sequential refinement.}
\end{figure}

English typically converges in 1–2 refinement steps. Urdu needs 2–4, signaling deeper baseline instability.

\subsection{English–Urdu Disparity Analysis}

Table 2 shows language deltas:
\begin{itemize}
    \item Baseline gap: 0.16625
    \item PRM-Select gap: 0.017 (almost eliminated)
    \item PRM-Sequential: 0.060 (gap reappears due to Urdu utility drop)
\end{itemize}

PRM-Select is the most cross-lingually fair method.  
Sequential refinement maximizes fairness but is less stable for Urdu.

\section{Discussion}

\subsection{Inference-Time Debiasing Works}

Both PRM-Select and PRM-Sequential significantly reduce bias across languages, extending existing evidence for preference-based inference-time safety mechanisms. These results demonstrate that alignment techniques need not require fine-tuning.

\subsection{Urdu's Baseline Disadvantage Reflects Structural Imbalances}

English dominates global pretraining corpora. This leads to:
\begin{itemize}
    \item cleaner embeddings,
    \item reduced hallucination,
    \item lower stereotype leakage,
    \item better semantic fit.
\end{itemize}

Urdu’s weaker baseline aligns with cross-lingual fairness studies showing performance is tightly correlated with resource availability (Blasi et al., 2022).

\subsection{PRM-Select vs.\ PRM-Sequential}

PRM-Sequential yields the highest fairness scores but sometimes harms utility, especially in Urdu. PRM-Select is more robust overall:

\begin{itemize}
    \item stable across both languages,
    \item computationally cheaper,
    \item nearly eliminates English–Urdu disparity.
\end{itemize}

Sequential refinement is preferable when maximal fairness is required, PRM-Select when consistency is a priority.

\subsection{Utility Behavior in Low-Resource Languages}

Urdu utility sometimes decreases because refinement steps nudge GPT-3.5 toward overly neutral or generic words. This supports critiques that debiasing can oversmooth language representations (Barocas and Selbst, 2017).

\subsection{Cross-Lingual Implications}

This work contributes to emerging research on multilingual fairness.  
Unlike prior work—mostly on embeddings or QA—this study targets **single-word LLM generation**, highlighting new cross-lingual asymmetries.

\section{Limitations}

\begin{enumerate}
    \item PRM zero-shot scoring may be inconsistent across contexts.
    \item Single-word generation simplifies fairness but limits generalization.
    \item Urdu translations cannot fully capture culturally nuanced stereotypes.
    \item No human annotator study was conducted.
\end{enumerate}

\section{Conclusion}

This paper introduces a unified inference-time debiasing pipeline and applies it to English and Urdu single-word generation tasks. Key findings:

\begin{itemize}
    \item Urdu is substantially more biased at baseline.
    \item PRM-Select provides the most balanced improvements.
    \item PRM-Sequential maximizes fairness at the cost of Urdu utility.
    \item Inference-time approaches can meaningfully reduce cross-lingual inequalities.
\end{itemize}

Future work may incorporate full-sentence debiasing, human preference tuning, or additional low-resource languages.

\section*{References}

S Ahmed and M Khan. \textit{Urdu word embeddings and linguistic bias}. Journal of Asian Language Technologies, 2019.

Amanda Askell et al. \textit{A general language assistant as a laboratory for alignment}. arXiv:2112.00861, 2022.

Solon Barocas and Andrew Selbst. \textit{Fairness in machine learning: A critical survey}. Foundations and Trends in Machine Learning, 2017.

Emily M. Bender, Timnit Gebru, Angelina McMillan-Major, and Margaret Mitchell. \textit{On the dangers of stochastic parrots}. FAccT, 2021.

Damian Blasi, Antonios Anastasopoulos, and Graham Neubig. \textit{Systematic inequalities in language technology performance}. PNAS, 2022.

Tolga Bolukbasi et al. \textit{Debiasing word embeddings}. NeurIPS, 2016.

Wei-Lin Chiang et al. \textit{Vicuna: An open-source chatbot}. arXiv:2306.0564, 2023.

Jacob Devlin et al. \textit{BERT}. NAACL, 2019.

Deep Ganguli et al. \textit{Red teaming LLMs}. arXiv:2209.07858, 2022.

Micah Goldblum et al. \textit{Inference-time attacks on LLMs}. arXiv:2309.06209, 2023.

Hila Gonen and Yoav Goldberg. \textit{Lipstick on a pig}. NAACL, 2019.

Dan Hendrycks et al. \textit{Measuring massive multitask language understanding}. ICLR, 2020.

Q. Huang et al. \textit{Generating unbiased texts}. EMNLP, 2020.

Nathan Lambert et al. \textit{Fairness in large language models}. arXiv:2305.13972, 2023.

Percy Liang et al. \textit{Holistic evaluation of language models}. TMLR, 2022.

Stephanie Lin et al. \textit{TruthfulQA}. arXiv:2109.07958, 2021.

S. Liu et al. \textit{Fairness in multilingual NLP}. ACL, 2022.

Ninareh Mehrabi et al. \textit{Survey on bias and fairness}. ACM Computing Surveys, 2021.

A. Mishra et al. \textit{Cross-lingual bias evaluation}. ACL, 2021.

Saif M. Mohammad. \textit{Gender norms and gendered language}. ACL, 2020.

Moin Nadeem et al. \textit{StereoSet}. ACL, 2021.

Nikita Nangia et al. \textit{CrowS-Pairs}. EMNLP, 2020.

Debora Nozza et al. \textit{Mitigating bias in LLMs}. ACL, 2023.

Long Ouyang et al. \textit{Training language models to follow instructions}. arXiv:2203.02155, 2022.

Alicia Parrish et al. \textit{BBQ}. ACL, 2022.

Alec Radford et al. \textit{Language models are unsupervised multitask learners}. OpenAI, 2019.

Colin Raffel et al. \textit{T5}. JMLR, 2020.

Pedro Salinas et al. \textit{Inference-time bias mitigation}. ACL, 2024.

Maarten Sap et al. \textit{Social bias frames}. ACL, 2020.

Timo Schick and Hinrich Schütze. \textit{Pattern-exploiting training}. ACL, 2021.

Emily Sheng et al. \textit{Biases in language generation}. EMNLP, 2019.

A. Srivastava et al. \textit{BLOOM}. arXiv:2211.05100, 2022.

Alex Tamkin et al. \textit{Capabilities and limitations of LLMs}. arXiv:2102.02503, 2021.

Hugo Touvron et al. \textit{LLaMA}. arXiv:2302.13971, 2023.

Miles Turpin et al. \textit{Language models don’t always say what they think}. NeurIPS, 2023.

Ben Wang. \textit{Self-debiasing LLMs}. arXiv:2303.01474, 2023.

Laura Weidinger et al. \textit{Ethical and social risks of harm from LLMs}. arXiv:2112.04359, 2021.

Guntur Winata et al. \textit{Neural machine translation and Urdu linguistic challenges}. TACL, 2023.

Guntur Winata et al. \textit{Multilingual fairness in language generation}. TACL, 2024.

Mengzhou Xia et al. \textit{Low-resource language models: A survey}. arXiv:2109.01247, 2021.

Jieyu Zhao et al. \textit{Men also like shopping: Reducing gender bias}. EMNLP, 2017.

Jieyu Zhao et al. \textit{Gender bias in coreference resolution}. NAACL, 2018.

W. Zhou et al. \textit{Survey of prompt engineering}. ACL, 2023.

\end{document}